\pdfoutput=1

\documentclass[11pt]{article}

\usepackage[]{ACL2023}

\usepackage{times}
\usepackage{latexsym}
\usepackage[T1]{fontenc}
\usepackage[utf8]{inputenc}
\usepackage{microtype}
\usepackage{inconsolata}
\usepackage[ruled,vlined]{algorithm2e}
\usepackage{hyperref}
\usepackage{url}
\usepackage{tcolorbox}
\usepackage{hyperref}
\usepackage{wrapfig}
\usepackage{subcaption}
\usepackage{pifont}
\usepackage{multirow}
\usepackage{verbatim}
\usepackage[skip=5pt]{caption} 

\usepackage{enumitem}
\usepackage{amsthm}
\usepackage{amsmath}
\usepackage{amssymb}
\usepackage{etoolbox}
\usepackage{booktabs}
\usepackage{float}
\usepackage{makecell}
\usepackage{acronym}

\usepackage{mathtools}
\usepackage{mathrsfs}

\acrodef{FT}{fine-tuning}
\acrodef{IF}{instruction following}
\acrodef{PEFT}{parameter-efficient fine-tuning}
\acrodef{LoRA}{low-rank adaptation}
\acrodef{MELoRA}{mini-ensemble low-rank adapters}
\acrodef{LLM}{large language model}
\acrodef{NLU}{natural language understanding}
\acrodef{NLP}{natural language processing}
\acrodef{SOTA}{state-of-the-art}

\newcommand{\model}{\acs{MELoRA}}

\title{MELoRA: Mini-Ensemble Low-Rank Adapters\\ for Parameter-Efficient Fine-Tuning}

\author{
    Pengjie Ren\textsuperscript{\rm 1}\thanks{~~~Equal contribution.}~~, \ 
    Chengshun Shi\textsuperscript{\rm 1}$^{*}$, \ 
    Shiguang Wu\textsuperscript{\rm 1}, \ 
    Mengqi Zhang\textsuperscript{\rm 1}, \ 
    Zhaochun Ren\textsuperscript{\rm 2},\\\ 
    \textbf{Maarten de Rijke\textsuperscript{\rm 3},}\ 
    \textbf{Zhumin Chen\textsuperscript{\rm 1},} \  
    \textbf{Jiahuan Pei\textsuperscript{\rm 4}\thanks{$^\dagger$ Corresponding author.}} \\
    \textsuperscript{\rm 1} Shandong University   \quad
    \textsuperscript{\rm 2} Leiden University  \\
    \textsuperscript{\rm 3} University of Amsterdam \quad
    \textsuperscript{\rm 4} Centrum Wiskunde \& Informatica \\
    \texttt{\{shichengshun,shiguang.wu\}@mail.sdu.edu.cn}, \\
    \texttt{\{renpengjie,mengqi.zhang,chenzhumin\}@sdu.edu.cn},\\
    \texttt{\ z.ren@liacs.leidenuniv.nl},
    \texttt{\ m.derijke@uva.nl},
    \texttt{\ jiahuan.pei@cwi.nl }
}
        
\begin{document}
\maketitle

\begin{abstract}

\Acf{PEFT} is a popular method for tailoring pre-trained \acp{LLM}, especially as the models' scale and the diversity of tasks increase.
\Ac{LoRA} is based on the idea that the adaptation process is intrinsically low-dimensional, i.e., significant model changes can be represented with relatively few parameters.
However, decreasing the rank encounters challenges with generalization errors for specific tasks when compared to full-parameter fine-tuning.
We present \acs{MELoRA}, a \acl{MELoRA} that uses fewer trainable parameters while maintaining a higher rank, thereby offering improved performance potential.
The core idea is to freeze original pretrained weights and train a group of mini \acp{LoRA} with only a small number of parameters.
This can capture a significant degree of diversity among mini \acp{LoRA}, thus promoting better generalization ability.
We conduct a theoretical analysis and empirical studies on various NLP tasks.
Our experimental results show that, compared to \ac{LoRA}, \acs{MELoRA} achieves better performance with 8 times fewer trainable parameters on natural language understanding tasks and 36 times fewer trainable parameters on instruction following tasks, which demonstrates the effectiveness of \acs{MELoRA}.

\end{abstract}
\section{Introduction}

\begin{figure}[t!]
\centering  
\includegraphics[width=0.5\textwidth]{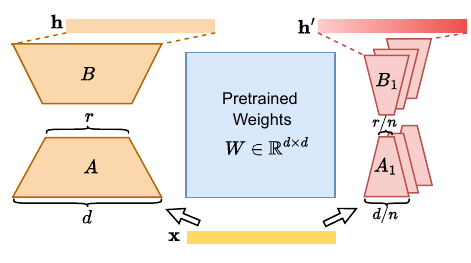}  
\caption{Comparison between \ac{LoRA} (left) and the proposed \model~ (right).
The core idea of \model\ is to freeze original pretrained weights and train a group of mini \acp{LoRA} in parallel with only a small number of parameters. 
}
\label{fg:model} 
\end{figure}

\Acfp{LLM} have emerged as the default paradigm for \ac{NLP}~\cite{gpt3}.
\Acf{FT} is a prevailing way for tailoring \acp{LLM} for specific downstream tasks~\cite{peft2023ding}.
However, as the models' scale and the diversity of the tasks increase, fully \ac{FT} becomes infeasible.
\Acf{PEFT} has been proposed to alleviate memory demands by reducing trainable parameters~\citep{adapter2,prefix-tuning,prompt-tuning,lora}.
Typically, the core idea of \ac{PEFT} methods is to update only a small fraction of the parameters, such as adapter weights~\cite{adapter2,lora} and prompt weights~\cite{prefix-tuning,prompt-tuning}.

\Acf{LoRA}~\citep{lora} is widely being used due to its minimal additional memory overhead and because it comes without  additional inference latency.
As illustrated in Figure~\ref{fg:model} (left),
\ac{LoRA} uses low-rank matrices ($A, B$) to approximate the updates of pre-trained weights ($W$).
As the rank is smaller than the model's hidden dimension, the overall number of trainable parameters of \ac{LoRA} is much smaller than full \ac{FT}.
Despite the significant computational advantage, low-rank approximation may lead to a substantial performance gap when compared to full \ac{FT}~\citep{lora,deltalora}.
Therefore, the following is a critical challenge:
\begin{quote}
\textit{How to enable a higher rank variation while preserving the computational advantage?}
\end{quote}
To increase the rank of \ac{LoRA} without introducing more trainable parameters, ReLoRA~\citep{lialin2023stack} and COLA~\citep{cola} append multiple \acp{LoRA} to pre-trained weights.
They progressively merge old \ac{LoRA} to pre-train weights and stack new \acp{LoRA} during training.
In essence, these methods train multiple \acp{LoRA} in series.
However, there may be overlap between the series of \ac{LoRA} modules.
Therefore, employing the direct sum of multiple \ac{LoRA} modules in these methods does not necessarily guarantee an increase in rank.
In this work, we propose a simple yet effective method, called \acf{MELoRA}, that stacks multiple mini \acp{LoRA} in parallel, as shown in Figure~\ref{fg:model} (right).
We demonstrate theoretically that \model{} ensures a higher rank without imposing an additional parameter overhead.
We concatenate multiple mini \acp{LoRA} simultaneously along the diagonal to construct an equivalent block diagonal \ac{LoRA} matrix.
Each mini \ac{LoRA} is therefore independent of the other and the final rank will be the sum of the rank of each mini LoRA.
Each mini \ac{LoRA} rank just learns the different dimensions of the hidden state.
The shape of the trainable weights $A$ and $B$ in mini \ac{LoRA} will be much thinner.

We conduct extensive experiments across diverse tasks and models to demonstrate the efficacy of \model.
Evaluations are performed using RoBERTa-base on natural language understanding tasks and Llama-2-7B on instruction following tasks.
Results indicate that \model{} achieves superior performance while using significantly fewer parameters. 
For instance, with 36 times fewer trainable parameters of LoRA, \model{} outperforms \ac{LoRA} on all instruction following datasets.

We summarize our contributions as follows:
\begin{itemize}[leftmargin=*,nosep] %
\item We propose a new method (\model) on top of \ac{LoRA} that makes it achieve a higher rank and better performance with fewer parameters.
\item We theoretically demonstrate that \model{} maintains a higher and flexible rank, as well as lower complexity, compared to \ac{LoRA}.
\item Extensive experiments show that \model{} outperforms \ac{LoRA} in terms of parameter quantity and performance.
\end{itemize}

\section{Related Work}

Full-parameter fine-tuning poses computational challenges with growing model sizes and the proliferation of downstream tasks. 
In response to these challenges, 
\acf{PEFT}, modifying only a small portion of parameters while leaving the majority of pre-trained model parameters unchanged, has received increasing attention from researchers.
Numerous studies \citep{adapter2, adapter1, adapter-fusion, adapter-drop, prefix-tuning, prompt-tuning, lora} discuss key factors such as reduced inference overhead, memory efficiency, and storage optimization.
In particular, LoRA \cite{lora} introduces trainable low-rank matrices to approximate weight updates during fine-tuning.
Due to its simplicity in implementation without inducing any noticeable latency during inference, it is widely used in many fields.

A number of advanced techniques have been proposed that build on the basic principles of LoRA.
These extensions fall into two main categories: adaptive rank and customized update strategies.

\subsection{Adaptive Rank} 
Some studies~\cite{adalora,search} argue that while the popular Low-Rank Adaptation (LoRA) method is effective, it uses a fixed intrinsic rank that may not always be optimal. 
They highlight the efficacy of employing higher ranks for more important parameters.
Notably, AdaLoRA \cite{adalora} adopts an adaptive approach to singular value pruning, tailoring rank selection based on the magnitude of individual singular values.
Consequently, this method involves the use of different ranks across various layers.
Similarly, \citet{sora} use a gate unit to facilitate the pruning of different ranks.
In contrast, \citet{IncreLoRA} propose IncreLoRA, an incremental parameter allocation method that adaptively adds trainable parameters during training based on the importance scores of each module.

\subsection{Customized Update Strategies}

Another way to improve LoRA is to change the parameter update strategy.
Some work is devoted to reducing the number of trainable parameters.
\citet{VeRA} introduces a shared frozen random LoRA module applicable to all pre-trained weights and only trains scaling vectors between LoRA $B$ and $A$ matrices to curtail the number of trainable parameters.
This approach reduces the trainable parameter count by a factor of 10 compared to conventional LoRA.
Another approach by \cite{lora-fa} involve freezing the $A$ matrix within LoRA, effectively halving the count of trainable parameters.
But both of the these methods incur a substantial drop in performance.

QLoRA \cite{qlora} further leverages 4-bit quantization to effectively and efficiently fine-tune LLMs.
LoRAMoE \cite{loramoe} uses multiple LoRAs as adaptable experts and a router to gate them in the feed-forward network layer to address the problem that fine-tuning data can disrupt the world knowledge stored in LLMs. 

To perform model inference in different rank settings, drawing inspiration from nested dropout techniques, \citet{dylora} propose a dynamic parameter update strategy, enabling a single training process for multiple rank inferences.
Delta-LoRA \cite{deltalora} updates not only the low-rank matrices $A$ and $B$ but also propagates the learning to the pre-trained weights $W$ using the delta of the product of two low-rank matrices $A$ and $B$. 

Despite the computational advantages, so far, low rank approximation leads to a substantial performance gap.
To address this limitation, ReLoRA~\citep{lialin2023stack} and COLA~\citep{cola} append multiple LoRAs to pre-trained weight to increase the rank of LoRA without introducing more trainable parameters.
They progressively merge old LoRA layers to pre-train weight and stack new LoRA layers during training.
However, there is no theoretical guarantee for the rank lower bound in training.

Unlike previous work, our proposed method concatenates multiple mini LoRAs in parallel along the diagonal to construct a block diagonal LoRA matrix.
It ensures that the final rank will be the sum of the ranks of each mini LoRA.

\section{Methodology}

In this section, we introduce the proposed \acf{MELoRA}, a novel method that involves concatenating the outputs from several mini LoRA modules, as illustrated in Figure~\ref{fg:model}.

\subsection{Preliminaries on Low-Rank Adapter} 
LoRA decomposes the weight update $\Delta W$ into a low-rank product $B A$.
During training, the pre-trained weights $W$ are frozen and do not receive gradient updates, while $A$ and $B$ contain trainable parameters as shown in Equation~\ref{eq:lora}:
\begin{equation}
h = W x + \Delta W x = W x + BA x,
\label{eq:lora}
\end{equation}
where 
$W \in \mathbb{R}^{d \times d}$,
$A \in \mathbb{R}^{r \times d}$,
$B \in \mathbb{R}^{d \times r}$,
$x \in \mathbb{R}^{d}$,
and $r \ll d$.
At the start of the training stage, $A$ is randomly initialized via Gaussian initialization, and $B$ is initialized to a zero matrix to make ensure the incremental update $BA = 0$ at initialization.

\subsection{Matrix Rank Theory} 
\label{sec:rank}

In linear algebra, several useful inequalities govern the rank of matrices:
\begin{align}
    \mathcal{R}(M_1 + M_2) & \le \mathcal{R}(M_1) + \mathcal{R}(M_2),
    \label{eq:rank_ineq1}
    \\
\begin{split}
    \max(\mathcal{R}(M_1), {}& \mathcal{R}(M_2)) \\
    &\le\;\mathcal{R}(\text{concat}(M_1, M_2)) \\
    &\le\;\mathcal{R}(M_1) + \mathcal{R}(M_2),
    \label{eq:rank_ineq2}
\end{split}
\\
    \mathcal{R}(\text{diag}_{i=0}^{n} M_i) & = \sum_{i=1}^{n} \mathcal{R}(M_i),
    \label{eq:rank_ineq3}
\end{align}
where $\mathcal{R}(\cdot)$ denotes the operation to get the rank of a matrix.
Equation~\ref{eq:rank_ineq1} demonstrates that there is no lower bound when matrices undergo simple addition operations. 
Equation~\ref{eq:rank_ineq2} indicates that the rank does not increase by concatenation, as the column vectors may exhibit linear correlations. 
However, when matrices are concatenated diagonally, as per Equation~\ref{eq:rank_ineq3}, the final rank becomes the sum of each matrix's rank.

\subsection{Mini-Ensemble Low-Rank Adapter} 
\label{sec:parameter}

In \model, we employ $n$ mini LoRAs on pre-trained weights, as depicted on the right in Figure~\ref{fg:model}.
For the convenience of comparison with LoRA, we set the rank of each mini LoRA to $\frac{r}{n}$.
\model{} is defined as the concatenation of several mini LoRAs across different hidden dimensions:
\begin{equation}
\begin{aligned}
h &= W x + \Delta W x \\
 &= W x + \left(\operatorname{concat}_{i=0}^{n}B_iA_i x_i\right) \\
 &= W x + \left(\operatorname{diag}_{i=0}^{n}B_iA_i\right) x \\
 & = W x + \left(\operatorname{diag}_{i=0}^{n}B_i \right) \left(\operatorname{diag}_{i=0}^{n}A_i\right) x,
\label{eq:ours}
\end{aligned}
\end{equation}
where 
$A_i \in \mathbb{R}^{\frac{r}{n} \times \frac{d}{n}}$, 
$B_i \in \mathbb{R}^{\frac{d}{n} \times \frac{r}{n}}$,  
$x, h \in \mathbb{R}^{d}$ 
, $x_i \in \mathbb{R}^{\frac{d}{n}}$ is the feature split from $x$,
and $n$ represents how many mini LoRA modules we use to concatenate.

\begin{figure}[t!]
\centering  
\includegraphics[width=0.5\textwidth]{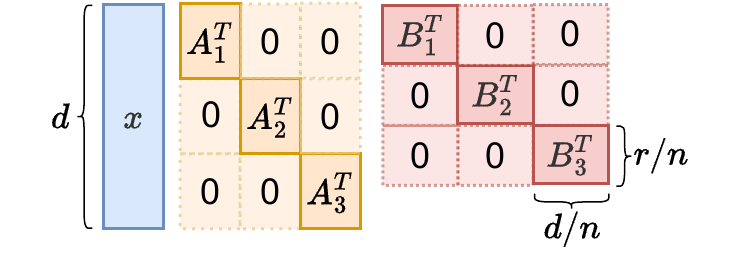}  
\caption{An illustration of how in \model{} adopt a group of mini LoRA modules to obtain sparse equivalent $B$, $A$.
    $x \in \mathbb{R}^{d}$ denote a representation with $d$ dimensions, 
    $A_i \in \mathbb{R}^{\frac{r}{n} \times \frac{d}{n}}$, $B_i \in \mathbb{R}^{\frac{d}{n} \times \frac{r}{n}}$ ($r \ll d)$, and 0 denotes zero metrics requiring no training.
}
\label{fg:sparce} 
\end{figure}

As deduced in Equation~\ref{eq:ours}, \model{} may be regarded as a form of sparse LoRA.
Figure~\ref{fg:sparce} illustrates the process of obtaining equivalent $B$ and $A$ matrices by padding zeros to $B_i$ and $A_i$ along the non-diagonal lines.
According to Equation~\ref{eq:rank_ineq3}, the rank of equivalent $B$, $A$ is the sum of individual ranks $B_i$ and $A_i$. 
Because $\frac{r}{n} \ll d$, each $B_i$ and $A_i$ possesses the same rank $\frac{r}{n}$; the resulting equivalent rank is $n \times \frac{r}{n} = r$.

We employ an identical initialization method to that of LoRA, wherein each $A_i$ undergoes random Gaussian initialization and $B_i$ is initialized to zero.

Compared to LoRA, \model{} has the following three advantages:

\textbf{(1) \model{} maintains a higher rank with fewer parameters.}
\label{rank}
The capability of \model{} to achieve a higher rank with fewer parameters is notable. 
As discussed in Section~\ref{sec:rank}, the simple summation or concatenation of matrices may not inherently increase rank due to potential overlaps between them. 
In \model, the matrices $B_i$ and $A_i$ are arranged in distinct columns and rows, ensuring that the rank of $\operatorname{diag}_{i=0}^{n}B_i$ and $\operatorname{diag}_{i=0}^{n}A_i$ is the sum of individual ranks $B_i$ and $A_i$.
Figure~\ref{fg:sparce} illustrates that the number of trainable parameters in \model{} is determined by the expression $ n \times (\frac{d_{\text{in}}}{n} \times \frac{r}{n} + \frac{r}{n} \times \frac{d_{\text{out}}}{n}) = \frac{d_{\text{out}} \times r + r \times d_{\text{in}}}{n} $. 
When achieving the same rank, the number of trainable parameters is $d_{\text{out}} \times r + r \times d_{\text{in}}$ \citep{lora} for LoRA. 
Importantly, the number of trainable parameters in \model{} is proportionally reduced by a factor of $n$ compared to LoRA. 
This suggests the potential for attaining a larger rank while utilizing fewer parameters within the \model{} framework.

\textbf{(2) \model{} has a more flexible rank.} 
The ability to alter the rank without necessitating changes in parameter count is another advantage of \model. 
Recent studies \citep{lora,dylora} emphasize the significance of rank variation across different datasets in influencing model performance. 
In \model, we might as well set the rank of each mini LoRA to $r$.
So individual mini LoRA modules denoted as $A$ in $\mathbb{R}^{r \times \frac{d}{n}}$ and $B$ in $\mathbb{R}^{\frac{d}{n} \times r}$ are configured, with a total count of trainable parameters being $2 \times r \times d$ and the equivalent rank expressed as $n \times r$. 
Adjusting the hyperparameter $n$ allows for modulation of the equivalent rank without necessitating an increase in the overall parameter count.

\textbf{(3) \model{} has lower complexity.}
We can compare the complexity of LoRA and \model{} under equal rank conditions, 
where $A \in \mathbb{R}^{r \times d}$, $B \in \mathbb{R}^{d\times r}$, $A_i \in \mathbb{R}^{\frac{r}{n} \times \frac{d}{n}}$, $B_i \in \mathbb{R}^{\frac{d}{n} \times \frac{r}{n}}$, and $x \in \mathbb{R}^{d}$.
The time complexity of LoRA is $ dr + rd=2rd$,
while each mini LoRA in \model{} is $\frac{d}{n}\times \frac{d}{n} + \frac{d}{n} \times \frac{d}{n} =\frac{2rd}{n^2}$. 
So the total operations of \model{} is $n \times \frac{2rd}{n^2} =\frac{2rd}{n}$.
Since each mini-LoRA module operates independently and can be computed in parallel, the overall complexity of \model{} is $\frac{2rd}{n^2}$.
Thus, while \model{} executes $n$ times fewer operations compared to LoRA, the final time complexity is reduced by a factor of $n^2$ relative to LoRA.

\section{Experimental Setups}

\subsection{Baselines}
We compare \model~with LoRA and a number of state-of-the-art LoRA variants:

\begin{itemize}[leftmargin=*,nosep]
    \item \textbf{LoRA}~\citep{lora} uses the multiplication of two low-rank matrices to learn the incremental updates with reduced GPU memory cost. 

    \item \textbf{DyLoRA}~\citep{dylora} randomly selects a rank $r$ for LoRA modules during learning. 

    \item \textbf{AdaLoRA}~\citep{adalora}  focuses on determining the optimal rank for incremental updates. It employs an adaptive approach to singular value pruning, tailoring the rank selection to the magnitude of each singular value. Consequently, distinct ranks are employed for different layers.

    \item \textbf{Delta-LoRA}~\citep{deltalora} not only updates the low-rank matrices $A$ and $B$ but also propagates the learning to the pre-trained weights $W$ via updates using the delta of the product of two low-rank matrices $(A^{(t+1)}B^{(t+1)} - A^{(t)}B^{(t)})$. 
    We follow their setups to reproduce NLU experimental results for a fair comparison.
\end{itemize}

\subsection{Datasets}
We evaluate the performance on two groups of datasets: GLUE \cite{glue} and INSTRUCTEVAL \cite{instructeval}.
The statistics are shown in Table~\ref{tab:dataset_stat}.
The GLUE benchmark is for NLU tasks~\cite{glue} and includes classification tasks, similarity and paraphrase tasks, and natural language inference tasks:

\begin{itemize}[leftmargin=*,nosep]
\item \textbf{MRPC} \cite{mrpc} is a corpus of sentence pairs automatically extracted from online news sources.
The task is to determine whether the sentences in a given pair are semantically equivalent. 
\item \textbf{RTE} comes from a series of annual textual entailment challenges, i.e., RTE1~\cite{rte1}, RTE2~\cite{rte2}, and RTE3~\cite{rte3}.
The task is to predict whether the premise entails the entailment or not.
\item \textbf{QQP} is a collection of question pairs from the community question-answering website Quora.\footnote{\url{https://huggingface.co/datasets/glue/viewer/qqp}} 
The task is to determine whether a pair of questions are semantically equivalent. 
\item \textbf{CoLA} \cite{cola-d} consists of English acceptability judgments drawn from books and journal articles on linguistic theory. 
The task is to predict whether it is a grammatically correct English sentence.
\item \textbf{STSB} \cite{stsb} is a collection of sentence pairs drawn from news headlines, video and image captions, and natural language inference data.
Each pair is human-annotated with a similarity score from 1 to 5. 
The task is to predict these scores.
\item \textbf{SST-2} \cite{sst2} consists of sentences from movie reviews and human annotations of their sentiment. The task is to predict the sentiment of a given sentence. 
\item \textbf{QNLI} \cite{qnli} is a question-answering dataset consisting of question-paragraph pairs.
GLUE converts the task into sentence pair classification by forming a pair between each question and each sentence in the corresponding context. 
The task is to determine whether the context sentence contains the answer to the question.
\item \textbf{MNLI} \cite{mnli} is a crowdsourced collection of sentence pairs with textual entailment annotations. 
Given a premise sentence and a hypothesis sentence, the task is to predict whether the premise entails the hypothesis (entailment), contradicts the hypothesis (contradiction), or neither (neutral).

\end{itemize}

\noindent
We train our model on the cleaned Alpaca dataset\footnote{\url{https://huggingface.co/datasets/yahma/alpaca-cleaned}} and evaluate the performance on INSTRUCTEVAL, an instruction following benchmark \cite{instructeval}:

\begin{itemize}[leftmargin=*, nosep]
\item \textbf{MMLU} \cite{mmlu} is designed to measure world knowledge and problem-solving ability in multiple subjects.
It evaluates models in zero-shot and few-shot settings, making it more challenging and closer to how humans are evaluated. 
The benchmark covers 57 subjects across STEM, humanities, social sciences, and other areas, ranging in difficulty from elementary to advanced professional levels.  
Each sample has 4 choices. 
Given the instruction, the task is to predict the correct answer.

\item \textbf{BBH} \cite{bbh} is a subset of 23 challenging tasks from the BIG-Bench benchmark, which focuses on tasks believed to be beyond the capabilities of current language models. 
It requires models to follow challenging instructions such as navigation, logical deduction, and fallacy detection. 

\item \textbf{DROP} \cite{drop} is a math-based reading comprehension task that requires a system to perform discrete reasoning over passages extracted from Wikipedia articles. 
To perform well on DROP, a system must resolve references in a question to suitable parts of the given passage, and perform discrete operations such as addition, counting, or sorting. 

\item \textbf{HumanEval (HEval)} \cite{humaneval} is a problem-solving benchmark used for evaluating large language models trained on code.
It consists of 164 original programming problems that assess language comprehension, algorithms, and simple mathematics, with some problems comparable to simple software interview questions.
\end{itemize}

\begin{table}[t]
	\centering
		\begin{tabular}{c@{~}llr rrr}
		  \toprule 
			\multicolumn{2}{c}{BM} & Corpus & \#Train & \#Dev & \#Test \\ 
            \midrule
            \multirow{8}{*}{\rotatebox[origin=c]{90}{GLUE}}& 
            & MRPC & 3.7k & 408 & 1.7k\\ 
		&	&RTE & 2.5k & 276 & 3k  \\ 
	&		&CoLA & 8.5k & 1k & 1k \\
	&		&STS-B & 7k &1.5k& 1.4k\\ 
	&	&	SST2 & 67k & 872 & 1.8k \\ 
	&	&	QQP & 364k & 40k & 391k\\
	&	&	QNLI &  108k &5.7k&5.7k\\ 
	&	&	MNLI &  393k& 20k & 20k\\ 
             \midrule
              \multirow{4}{*}{\rotatebox[origin=c]{90}{INSTRUCT}} & \multirow{4}{*}{\rotatebox[origin=c]{90}{EVAL}} &MMLU &  99.8k & 1.8k & 14k \\ 
               & &BBH &  0 & 0 & 6.5k \\ 
               & &DROP &  77.4k & 0 & 9.5k \\ 
               & &HEval &  0 & 0 & 164 \\
            \bottomrule
		\end{tabular}
  	\caption{Dataset statistics. ``BM'' is short for ``Benchmark''. Only the test sets of INSTRUCTEVAL are used. We train the models on the cleaned Alpaca dataset.
   }
        \label{tab:dataset_stat} 
\end{table}

\subsection{Implementation Details}
For all experiments, we only fine-tune $W_Q$ and $W_V$~\cite{adalora}. 
All models and datasets are downloaded from Huggingface.\footnote{\url{https://huggingface.co}}
All models are fine-tuned on NVIDIA A800 GPUs.
The results are averaged with 5 different random seeds. 

\begin{table*}
  \centering
  \setlength{\tabcolsep}{5pt}
  \begin{tabular}{l r cccccccc c}
  \toprule
   Method & \#Params & MRPC & RTE & CoLA & STS-B & SST-2 & QQP & QNLI & MNLI & Avg. \\
  \midrule
  \acs{FT}$\ast$  & 125.0M & 88.23 & 84.11 & 64.57 & 90.56 & 94.26 & 91.96 & 92.73 & 87.51 & 86.74 \\ 
  \midrule
   DyLoRA$\ast$  & 295k & 89.46 & 84.47 & 61.12 & 91.06 & 94.26 & 90.17 & 92.22 & 86.33 & 86.14 \\ 
   AdaLora$\ast$  & 295k & 90.19 & 85.19 & 61.64 & 91.16 & 94.49 & 90.14 & 93.08 & 87.34 & 86.65 \\
   DeltaLoRA$\ast$  & 295k & 90.19 & \textbf{87.00} & 63.82 & \underline{91.57} & \underline{95.06} & \textbf{90.87} & \underline{93.09} & \textbf{87.50} & \underline{87.38} \\ 
   \midrule
   LoRA & 295k & 89.92 & 85.92 & 62.43 & 91.36 & 94.38 & \underline{90.78} & 92.64 & 86.91 & 86.79 \\
   \model~& 37k & \underline{90.69} & 86.28 & \underline{64.07} & 91.08 & 94.95 & 89.26 & 92.70 & 86.21 & 86.91\\ 
   \model~& 295k & \textbf{90.93} & \underline{86.64} & \textbf{64.09} & \textbf{91.93} & \textbf{95.41} & 90.77 & \textbf{93.17} & \underline{87.20} & \textbf{87.52} \\ 
  \bottomrule
  \end{tabular}
   \caption{
   Results on GLUE for natural language understanding tasks. 
   We report the overall (matched and mismatched) accuracy for MNLI, Matthew's correlation for CoLA, Pearson correlation for STS-B, and accuracy for other tasks. Higher is better for all metrics. 
  We also report the number of trainable parameters (\#Params) for each method.
  $\ast$ indicates the numbers published in \citep{deltalora}. 
  We use the same hyper-parameters as \citet{deltalora}.
  Boldface indicates the best results in terms of the corresponding metrics; the second-best results are \underline{underlined}.
  }
  \label{tab:NLU_main}
\end{table*}

On the GLUE benchmark,
we use RoBERTa-base with as the backbone \ac{LLM}.
For fair comparison, the training configurations are selected according to \citet{deltalora}.
We set the rank of LoRA and its variants to 8.
For \model, we performed experiments with two settings.
First, we set the rank of each mini LoRA to 8 in \model{} to get the same number of trainable parameters. 
Second, to assess performance with fewer trainable parameters, we conduct experiments with a rank of 1 for each mini LoRA.
As the number of trainable parameters of \model\ remains constant regardless of the number of mini LoRAs when the rank of mini LoRA is fixed, we explore the parameter $n$ from the set \{2, 4, 8\} and report the best performance.
And we also analyze the effect of $n$ in Section~\ref{sec:n}.

On the INSTRUCTEVAL benchmark, 
we use LLaMA-2-7B as the backbone \ac{LLM}, Alpaca dataset as train set, randomly select 2k samples as the development set.
Following INSTRUCTEVAL~\cite{instructeval}, we use 5-shot direct prompting for MMLU, 3-shot direct prompting for BBH, 3-shot direct prompting for DROP (dev), and 0-shot direct prompting for HEval.
During training, we use AdamW as the optimizer and train the models for 3 epochs. 
For fair comparison, we keep the number of epochs consistent with the baselines.
A linear learning rate schedule is applied with initial learning rate $3 \times 10^{-4}$.
The batch size is set to 128. 
We explore the rank of LoRA from the set \{8, 16, 32, 64, 128, 256\} and report the optimal performance.
For our method, \model, we set the rank $r$ to 1, explore the number of mini LoRAs $n$ from the set \{8, 16, 32, 64\}, and report the optimal performance.
More implementation details can be found in Appendix~\ref{sec:hyperparameter}.

\section{Results}
\label{sec:result}

\subsection{Performance on GLUE}

The results of all methods on GLUE are shown in Table~\ref{tab:NLU_main}.

We can see that \model\ outperforms LoRA on 7 out of 8 GLUE datasets under the same parameter setting.
Even using 8 times fewer parameters, \model{} still achieves better performance on 5 out of 8 datasets, underscoring the enhanced expressiveness and higher rank of \model.
It is worth noting that large improvements are achieved on MRPC, RTE, CoLA, and SST-2, which have limited training data.
We think the reason is that \model{} concatenates several mini LoRAs, which makes it more robust and has better generalization capability.
\model{} also achieves decent performance on the remaining datasets, including MNLI, QNLI and STS-B, which proves that \model{} is stable and reliable across different settings.

\subsection{Performance on INSTRUCTEVAL}

The results of all methods on INSTRUCTEVAL are shown in Table~\ref{tab:instrction_main}. 

\begin{table}[!ht]
  \centering
 \footnotesize
  \setlength{\tabcolsep}{1.5pt}
  \resizebox{\linewidth}{!}{
  \begin{tabular}{@{}l@{}rcccc@{}}
      \toprule
       Method & \#Params & MMLU & DROP & HEval & BBH\\
      \midrule 
      w/o FT & -  & 45.96 & 31.55 & 12.20 & 32.04 \\
      FT    & 7B & \textbf{47.30}	& 29.12	& 12.80	& 32.72 \\
    \midrule
      LoRA  & 33.6M & 45.64 & 32.46 & 15.09 & 32.40 \\
      QLoRA	& 33.6M	& 45.40	& 28.97	& 15.24	& 32.81 \\
      AdaLoRA & 33.6M & 45.96 & 31.94 & 14.02 & 32.85 \\
      \model~& 0.5M & 46.46 & \textbf{32.65} & \textbf{16.16} & \textbf{33.01} \\
      \bottomrule
      \end{tabular}
}
  \caption{ 
    Results on INSTRUCTEVAL for instruction-following tasks.
    We report the exact match for MMLU, DROP and BBH, pass@1 for HumanEval. 
    Higher is better for all metrics. 
    The boldface indicates the best results in terms of the corresponding metrics.
  }
  \label{tab:instrction_main}
\end{table}

We can see that \model{} consistently outperforms all baselines across all tasks while utilizing more than 36 times fewer trainable parameters.
This highlights the effectiveness and efficiency of our proposed approach in instruction following tasks.
As shown in Section \ref{sec:parameter}, we believe the reason is that \model{} can achieve a higher rank with fewer parameters.

\begin{table*}
  \centering
  \addtolength{\tabcolsep}{4pt}
  \begin{tabular}{l r r cccccccc c}
  \toprule
   Method & $r\times n$ & \#Param. & RTE & CoLA & STS-B & SST-2 & QNLI & Avg. \\
   \midrule
   LoRA & $8\times 1$& 295k & 75.63 & 62.34 & \textbf{90.71} & \textbf{94.50} & \textbf{92.55} & 83.14 \\
   \model~ & $4\times 2$& 147k & \textbf{76.39} & \textbf{62.70} & \textbf{90.71} & 94.33 & 92.52 & \textbf{83.33} \\
   \model~ & $2\times 4$ & \phantom{0}73k & 75.09 & 61.84 & 90.60 & 94.11 & 92.47 & 82.82 \\
  \midrule
   LoRA & $16 \times 1$ & 590k & 75.45 & \textbf{63.28} & 90.81 & \textbf{94.70} & 92.50 & 83.35 \\
   \model~ & $8\times 2$ & 295k & \textbf{75.93} & 63.10 & \textbf{90.82} & 94.61 & \textbf{92.61} & \textbf{83.42} \\
   \model~  & $4\times 4$ & 147k & 74.37 & 61.72 & 90.63 & 94.54 & 92.67 & 82.79 \\
    \model~ & $2\times 8$ & \phantom{0}73k & 73.65 & 61.36 & 90.41 & 94.18 & 92.41 & 82.40 \\
  \bottomrule
  \end{tabular}
  \caption{Performance on GLUE for natural language understanding tasks with different numbers of equivalent ranks ($r \times n$), with the same metrics as in Table~\ref{tab:NLU_main}. 
  Boldface indicates best results in terms of the corresponding metrics.
  }
  \label{tab:NLU_rank}
\end{table*}

\begin{table}[ht]
  \centering
  \footnotesize
  \addtolength{\tabcolsep}{-3pt}
  \begin{tabular}{l @{~} r @{~} r c c c c c}
  \toprule
   Method & \multicolumn{1}{c}{$r\times n$} & \#Param. & MMLU & DROP & BBH & Avg.\\
    \midrule 
    LoRA & $16 \times 1$ & 8.4M & 45.52 & 32.14 & 32.67 & 36.78\\
    \model~ & $8 \times 2$ & 4.2M & 45.63 & \textbf{32.97} & 32.61 & 37.07\\
    \model~ & $4 \times 4$ & 2.0M & 45.23 & 32.43 & 32.70 & 36.79\\
    \model~ & $2 \times 8$ & 1.0M & \textbf{46.53} & \textbf{32.97} & \textbf{33.06} & \textbf{37.52}\\
    \model~ & $1 \times 16$ & 0.5M & 45.38 & 31.52 & 33.29 & 36.73\\
    \midrule 
    LoRA & $32 \times 1$ & 16.8M & 45.30 & 32.33 & 32.42 & 36.68\\
    \model~ & $16 \times 2$ & 8.4M & 45.92 & 32.60 & 32.78 & 37.10\\
    \model~ & $8 \times 4$ & 4.2M & 46.05 & 32.16 & 33.09 & 37.10\\
    \model~ & $4 \times 8$ & 2.0M & \textbf{46.20} & \textbf{33.30} & 33.11 & \textbf{37.54}\\
    \model~ & $2 \times 16$ & 1.0M & 45.66 & 31.84 & 32.36 & 36.62\\
    \model~ & $1 \times 32$ & 0.5M & 45.60 & 31.55 & \textbf{33.35} & 36.83\\
    \midrule 
    LoRA & $64 \times 1$ & 33.6M & 45.66 & 32.46 & 32.43 & 36.85\\
    \model~ & $32 \times 2$ & 16.8M & 46.03 & \textbf{32.76} & 32.58 & 37.12\\
    \model~ & $16 \times 4$ & 8.4M & 46.15 & 32.68 & \textbf{33.11} & \textbf{37.31}\\
    \model~ & $8 \times 8$ & 4.2M & \textbf{46.43} & 32.57 & 32.55 & 37.18\\
    \model~ & $4 \times 16$ & 2.0M & 46.26 & 32.57 & 32.67 & 37.17\\
    \model~ & $2 \times 32$ & 1.0M & 45.43 & 32.41 & 32.93 & 36.92\\
    \model~ & $1 \times 64$ & 0.5M & 46.20 & 31.66 & 32.45 & 36.77\\
  \bottomrule
  \end{tabular}
  \caption{ 
      Performance on INSTRUCTEVAL for instruction following tasks with different equivalent ranks ($r \times n$), with the same metrics as in Table~\ref{tab:instrction_main}.
      Boldface indicates best results in terms of the corresponding metrics.   
      More results can be found in Appendix~\ref{sec:appendix_nr}.
  }
  \label{tab:instrction_rank}
\end{table}
\section{Analysis}

In this section, we analyze two key hyper-parameters in \model: the number of mini LoRAs $n$ and the rank of each mini LoRA $r$. 
According to Equation~\ref{eq:ours}, the equivalent rank of \model\ is denoted as \( n \times r\).
We investigate the effect of the equivalent rank in Section~\ref{sec:nr}, and analyze $n$ and $r$ separately in Section~\ref{sec:n} and~\ref{sec:r}. 

\subsection{Analysis of Equivalent Rank}
\label{sec:nr}
In this section, we delve into the effect of the equivalent rank.
We conduct experiments across different equivalent ranks, specifically 4, 8, and 16 on GLUE, and 16, 32, and 64 on INSTRUCTEVAL.
The results are shown in Table~\ref{tab:NLU_rank} and~\ref{tab:instrction_rank}, respectively.
 
We have two observations from the results.
First, \model\ consistently achieves superior or comparable performance across all equivalent rank settings.
In Table~\ref{tab:NLU_rank} and Table~\ref{tab:instrction_rank}, \model{} achieves the best performance on most datasets with more than 2 times fewer trainable parameters.
This indicates that equivalent rank is more important than the number of trainable parameters.
Ideally, we should search the best equivalent rank settings on different datasets, but setting $n$ to 2 is a good choice in most cases.
Second, the optimal equivalent ranks vary across datasets and tasks. 
On GLUE, the optimal performance is generally achieved with $n=2$. 
In contrast, on INSTRUCTEVAL, a higher value of $n$ such as 4 or 8 is a more effective choice.
We think that model sizes and task complexity are the main factors. 
For instance, the RoBERTa model, with only 125M parameters, is considerably smaller than Llama-2-7B.
According to scaling laws~\cite{scalinglaws}, Llama-2-7B is more powerful.
Consequently, larger models like Llama-2-7B may not necessitate a significant increase in trainable parameters to adapt, suggesting that \model\ plays a more pivotal role in larger models.

\begin{figure}[!ht]
    \centering
    \includegraphics[width=0.8\linewidth]{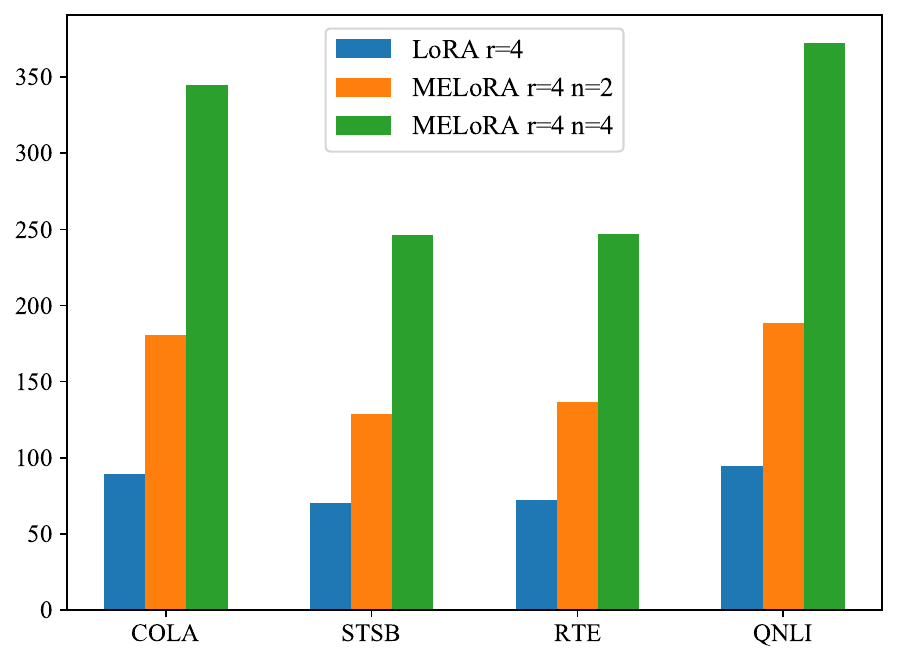}
    \caption{The sum of singular values $> 0.1$ of $B\times A$ in LoRA and equivalent $B\times A$ in \model. 
    }
    \label{fig:analysis_n_r}
\end{figure}

To ascertain whether \model~performs a higher rank update than LoRA, we analyze the number of singular values exceeding 0.1 for both LoRA and \model\ to estimate the real rank. 
As shown in Figure~\ref{fig:analysis_n_r}, \model~exhibits a significantly higher count of singular values compared to LoRA in the $n \times r$ setting.
This observation suggests that \model\ indeed achieves a high-rank update by performing multiple low-rank updates.

\begin{figure}[!ht]
    \centering
    \includegraphics[width=\linewidth]{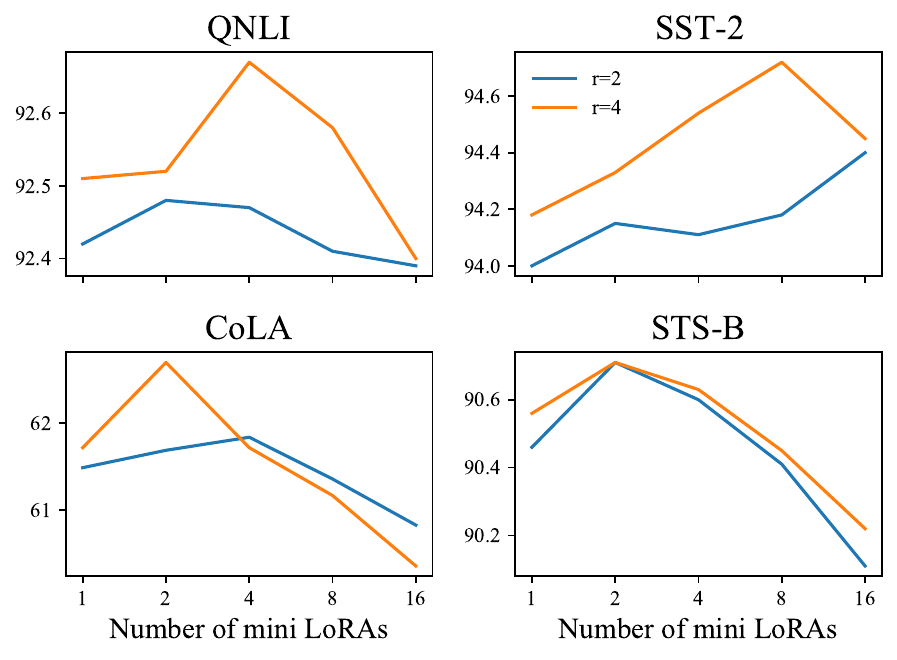}
    \caption{Performance with different number of mini LoRAs $n$ and fixed rank $r$ on different datasets.    
    We report the same metrics as Table~\ref{tab:NLU_main}.
    More results can be found in Appendix~\ref{sec:appendix_n}.}
    \label{fig:analysis_n}
\end{figure}

\subsection{Analysis of the Number of Mini LoRAs}
\label{sec:n}
As discussed in Section~\ref{sec:parameter}, maintaining a fixed rank for each mini LoRA results in an unaltered parameter count. 
Consequently, the ability to modify the equivalent rank by adjusting $n$ does not necessitate an increase in the overall number of parameters. 
To analyze the effect of $n$, we conduct experiments by varying $n$ while keeping $r$ fixed at 2 and 4 on four datasets. 
The results are shown in Figure~\ref{fig:analysis_n}.

We have three observations from the results.
First, the optimal $n$ varies across datasets and sometimes differs for different values of $r$ even on the same dataset. 
For instance, when $r$ is set to 4, the optimal $n$ for QNLI and SST-2 is 4, while for ColA and STSB, it is 2. 
This observation suggests that the specific task exerts an important influence on the behavior of the model.
Second, the performance of \model\ exhibits a pattern of initially increasing with $n$ and then decreasing, regardless of the values of $r$. 
At first, increasing $n$ results in a higher equivalent rank, which is beneficial to the performance.
However, note that excessively high equivalent ranks pose a risk of overfitting.
That is why when $n$ is too big, the performance drops.
Third, the optimal $n$ tends to be larger for datasets with more training samples or smaller values of $r$. 
As to training samples, for instance, QNLI and SST-2 have more training samples compared to other datasets, thus leading to an optimal $n$ of 4, while for the others, it is 2.
This phenomenon can be attributed to the need to allocate a higher equivalent rank to effectively leverage the abundance of training samples.
As to rank $r$, on SST-2 and CoLA, the optimal $n$ for $r=2$ is greater than that for $r=4$.
That is because the equivalent rank of \model\ is denoted as $n \times r$.
To achieve a specific equivalent rank, the smaller the value of $r$ is, the larger the value of $n$ should be.

\subsection{Analysis of the Rank of Mini LoRAs}
\label{sec:r}

\begin{figure}[!ht]
    \centering
    \includegraphics[width=\linewidth]{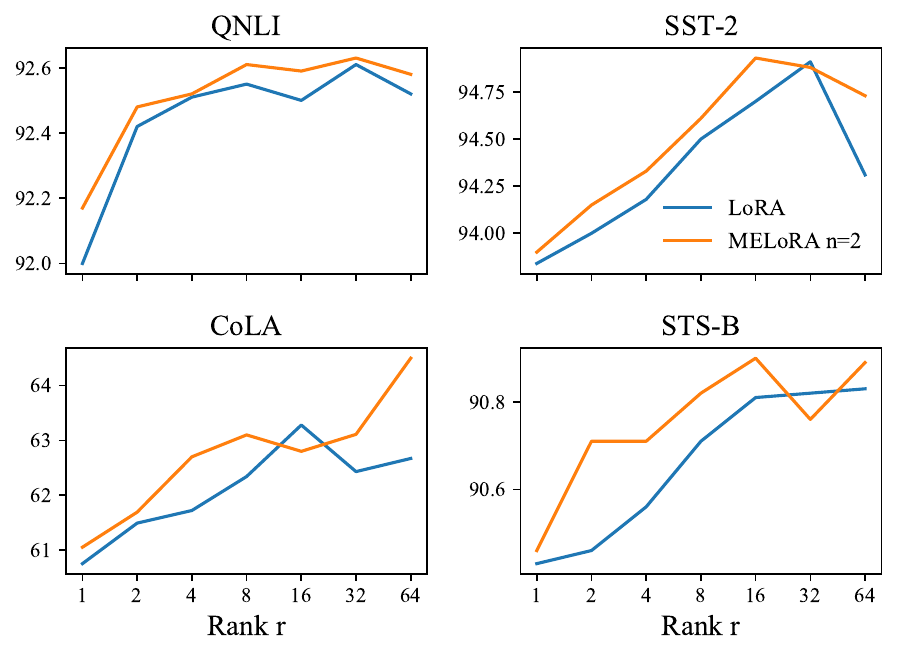}
    \caption{Performance of LoRA and \model\ with different rank $r$ and fixed $n$ on different datasets.
    More results can be found in Appendix~\ref{sec:appendix_r}.
    }
    \label{fig:analysis_r}
\end{figure}

To analyze the effect of $r$, we conduct experiments with varying $r$ while keeping $n$ fixed at 1 and 2 on four datasets.
When setting $n$ to 1, \model{} degrades into LoRA.
The results are shown in Figure~\ref{fig:analysis_r}.
As $r$ increases, the performance first improves and then stabilizes. 
This observation implies that a higher rank and a large number of trainable parameters are always favored in terms of performance when there is enough training data. 
However,  a higher rank and a large number of trainable parameters usually mean higher training costs.
Second, \model\ consistently outperforms LoRA across all rank settings.
That proves that \model\ is more powerful with the same number of trainable parameters.

\section{Conclusion}

In this paper, we have proposed a new parameter-efficient fine-tuning method, \model, that stacks multiple mini LoRAs in parallel, with each mini LoRA rank learning the different dimensions of a hidden state.
We have theoretically demonstrated that \model\ maintains a higher and flexible rank, as well as lower complexity.
We have also shown empirically that \model\ achieves a higher rank and better performance with fewer trainable parameters on multiple datasets.

\section*{Reproducibility}
We release the source code to reproduce the results in this paper at \url{https://github.com/ChasonShi/MELoRA}.

\section*{Limitations}
This work has the following limitations. 
First, we introduce a new hyper-parameter $n$, which indicates the number of mini-LoRAs.
The best $n$ is varied with different datasets.
We need more tuning parameters to get a good performance.
In future work, we plan to address these issues by applying hyper-parameter search methods like Bayesian optimization.

\section*{Ethical Considerations}
We realize that there are risks in developing large language models, so it is necessary to pay attention to the ethical issues.
We have used the public pre-trained \acp{LLM}, e.g., LLaMA2-7B, RoBERTa-base, and public datasets, i.e., GLUE and INSTRUCTEVAL, to conduct the experiments.
All models and datasets are carefully processed by their publishers to ensure that there are no ethical problems.

\section*{Acknowledgements}
This work was supported by the Natural Science Foundation of China (62102234, 62372275, 62272274, 62202271, T2293773, 62072279), the National Key R\&D Program of China with grant No.2022YFC3303004, the Natural Science Foundation of Shandong Province (ZR2021QF129), and by the Dutch Research Council (NWO), under project numbers 024.004.022, NWA.1389.20.\-183, and KICH3.LTP.20.006, and the European Union's Horizon Europe program under grant agreement No 101070212.
All content represents the opinion of the authors, which is not necessarily shared or endorsed by their respective employers and/or sponsors.

\bibliographystyle{acl_natbib}
\bibliography{references}

\clearpage

\appendix

\section{Hyper-parameters}
\label{sec:hyperparameter}

The detailed hyper-parameter settings on the INSTRUCTEVAL and GLUE datasets are listed in Table \ref{tab:hyper_parameter_instruction_train} and \ref{tab:hyper_parameter_glue}, respectively.

 \begin{table}[ht]
\centering
\setlength{\tabcolsep}{4mm}{
\begin{tabular}{lc}
\toprule  
Hyper-Parameter &  
\\
\midrule
    Learning rate $\eta$ & 3e-4  \\
    Batch size& 128 \\
    Number of epochs& 3\\
    Max sequence length & 256 \\
    Rank $r$ & 4\\
    LoRA dropout  & 0.05 \\
    LoRA alpha $\alpha$ & 16\\
    Trainable matrices & $W_Q$, $W_V$ \\
    LR scheduler & Linear \\
    Warmup steps & 100\\
    \bottomrule
    \end{tabular}
}
\caption{The hyper-parameter settings for INSTRUCTEVAL. We use the same settings as \citet{instructeval}}
\label{tab:hyper_parameter_instruction_train}
\end{table}

\begin{table*}[ht]
\centering

\setlength{\tabcolsep}{0.95mm}{
\begin{tabular}{l cccccccc}
\toprule Hyper-Parameter & MNLI & SST-2 & MRPC &CoLA & QNLI & QQP & RTE & STS-B \\
\midrule
Learning Rate $\eta$ &  5e-4 &  5e-4 &  4e-4 &  4e-4 &  4e-4 &  4e-4  &  4e-4  &  4e-4 \\
Batch Size &  128 &  128 &  128 &  64 &  128 & 128  &  128  &  128 
\\
Number of Epochs & 30 & 60 & 30 & 80 & 25 & 25 & 80 & 40 \\
Weight Decay $\beta$ & 0.1 & 0.1 & 0.1 & 0.1 & 0.1 & 0.1 & 0.1 & 0.1 \\
Max Sequence Length &  256 &  256 & 256 & 256 & 256 & 256 & 512 & 256 \\
Start Steps $K$ & 2000 & 400 & 10 & 100 & 800 & 400 & 200 & 200 \\
Update Ratio $\lambda$ & 0.5 & 0.5 & 0.5 &  0.5 & 0.5 & 0.5 & 0.5 & 0.5 \\
Rank $r$ & 8 & 8 & 8 & 8 & 8 & 8 & 8 & 8 \\
Alpha $\alpha$ & 16 & 16 & 16 & 16 & 16 & 16 & 16 & 16 \\
LR Scheduler & Linear & Linear & Linear & Linear & Linear & Linear & Linear & Linear \\
Trainable Matrices & $W_Q$,$W_V$ & $W_Q$,$W_V$ & $W_Q$,$W_V$ & $W_Q$,$W_V$ & $W_Q$,$W_V$ & $W_Q$,$W_V$ & $W_Q$,$W_V$ & $W_Q$,$W_V$ \\
Warmup Ratio & 0.06 & 0.06 & 0.06 & 0.06 & 0.06 & 0.06 & 0.06 & 0.06 \\
\multirow{2}{*}{Evaluation Metrics} & \multirow{2}{*}{Accuracy} & \multirow{2}{*}{Accuracy}  &\multirow{2}{*}{Accuracy}&Matthews&\multirow{2}{*}{Accuracy}&\multirow{2}{*}{Accuracy}&\multirow{2}{*}{Accuracy}&\multirow{2}{*}{Pearson} \\
& &&&Correlation&&&& \\
\bottomrule
\end{tabular}
}
\caption{The hyper-parameter settings for GLUE. For fair comparison, we use the same settings as \citet{deltalora}.}
\label{tab:hyper_parameter_glue}
\end{table*}

\section{Analysis of Equivalent Rank}
\label{sec:appendix_nr}
We further conduct experiments with equivalent ranks 128, 256 and 4096 on INSTRUCTEVAL.
The results are listed in Table~\ref{tab:instrct appendix}.
\model\ still achieves the best performance on all equivalent rank settings.
We have similar observations as in Table~\ref{tab:instrction_rank}.

\begin{table*}[!ht]
  \centering
  \begin{tabular}{l r r ccccc}
  \toprule
   Method & $r\times n$ & \#Param. & MMLU & DROP & HEval & BBH & Avg.\\
    \midrule 
      LoRA & $128\times 1$ & 67.1M & 45.36 & 32.52 & 14.63 & 32.32 & 31.20 \\
  \model & $64\times 2$ & 33.6M & 46.05 & 32.59 & 15.85 & 32.54 & 31.76\\     
  \model & $32\times 4$ & 16.8M & 46.05 & \textbf{32.78} & 15.24 & \textbf{33.18}  & 31.81 \\ 
  \model & $16\times 8$ & 8.4M & \textbf{46.40} & 32.49 & \textbf{16.46} & 32.85 & \textbf{32.05} \\
  \model & $8\times 16$ & 4.2M & 46.08 & 32.57 & 15.24 & 32.40 & 31.57 \\
  \model & $4\times 32$ & 2.0M & 45.82 & 32.38 & 15.85 & 32.37 & 31.61\\     
  \model & $2\times 64$ & 1.0M & 45.54 & 31.49 & 12.80 & 32.78 & 30.65 \\ 
  \model & $1\times 128$ & 0.5M & 45.71 & 31.69 &  14.02 & 32.20 & 30.91 \\
  \midrule
  LoRA & $256\times 1$ & 134.2M & 45.27 & 32.28 & \textbf{16.46} & 31.86 & 31.47  \\
  \model & $128\times 2$ & 67.1M & 45.95 & 32.73 & \textbf{16.46} & 32.51 & 31.91\\
  \model & $64\times 4$ & 33.6M & 45.94 & 32.95 & 15.85 & \textbf{33.25} & \textbf{32.00} \\
  \model & $32\times 8$ & 16.8M & \textbf{46.33} & \textbf{32.98} & 15.24 & 32.98 & 31.88 \\
  \model & $16\times 16$ & 8.4M & 46.26 & 32.73 & 14.02 & 32.30 & 31.33 \\
  \model & $8\times 32$ & 4.2M & 46.12 & 32.44 & 14.63 & 32.79 & 31.50 \\
  \model & $4\times 64$ & 2.0M & 46.28 & 31.31 & 12.80 & 32.46 & 30.71\\
  \model & $2\times 128$ & 1.0M & 45.40 & 32.04 & 14.63 & 31.74 & 30.95 \\
  \model & $1\times 256$ & 0.5M & 45.25 & 31.60 & 14.02 & 32.65 & 30.88 \\
  \midrule  
  \model & $2048\times 2$ & 1073.7M & 45.76 & 32.76 & 17.07 & 32.62 & 32.05 \\
  \model & $1024\times 4$ & 536.9M & 45.80 & 32.82 & \textbf{18.29} & 32.93 & \textbf{32.46}  \\  
  \model & $512\times 8$ & 268.4M & 46.35 & \textbf{32.89} & 15.24 & 32.82 & 31.83 \\  
  \model & $256\times 16$ & 134.2M & 46.11 & 32.51 & 15.85 & 32.52 & 31.75  \\
  \model & $128\times 32$ & 67.1M & 46.10 & 32.60 & 13.41 & 32.91 & 31.26  \\
  \model & $64\times 64$ & 33.6M & 45.96 & 32.04 & 15.85 & 32.14 & 31.49 \\
  \model & $32\times 128$ & 16.8M & 46.33 & 31.85 & 12.80 & 32.55 & 30.88 \\
  \model & $16 \times 256 $ & 8.4M & \textbf{46.45} & 32.30 & 13.41 & \textbf{32.79}  & 31.24 \\
  \model & $8\times 512$ & 4.2M & 46.40 & 32.09 & 14.63 & 32.97 & 31.52 \\
  \model & $4\times 1024$ & 2.1M & \textbf{46.45} & 32.12 & 14.02 & 32.87 & 31.37 \\
  \model & $2\times 2048$ & 1.0M & 45.99 & 32.33 & 13.41 & 32.41 & 31.04 \\ 
  \bottomrule
  \end{tabular}
  \caption{ 
  Performance on INSTRUCTEVAL for instruction following tasks with different equivalent ranks ($r \times n$), with the same metrics as in Table~\ref{tab:instrction_main}.
      Boldface indicates best results in terms of the corresponding metrics.   
  }
  \label{tab:instrct appendix}
\end{table*}

\section{Analysis of the Number of mini \acp{LoRA}}
\label{sec:appendix_n}

\begin{figure}[ht]
    \centering
    \includegraphics[width=\linewidth]{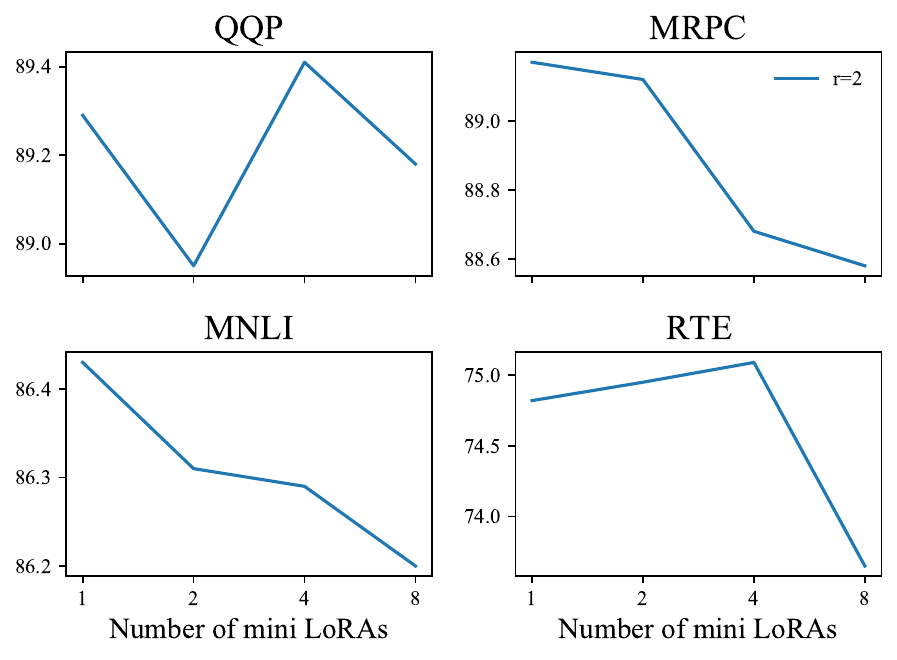}
    \caption{Performance of LoRA and \model\ with different number of mini LoRAs $n$ and fixed $r$ on GLUE.
    }
    \label{fig:analysis_n_appendix}
\end{figure}

\begin{figure}[ht]
    \centering
    \includegraphics[width=\linewidth]{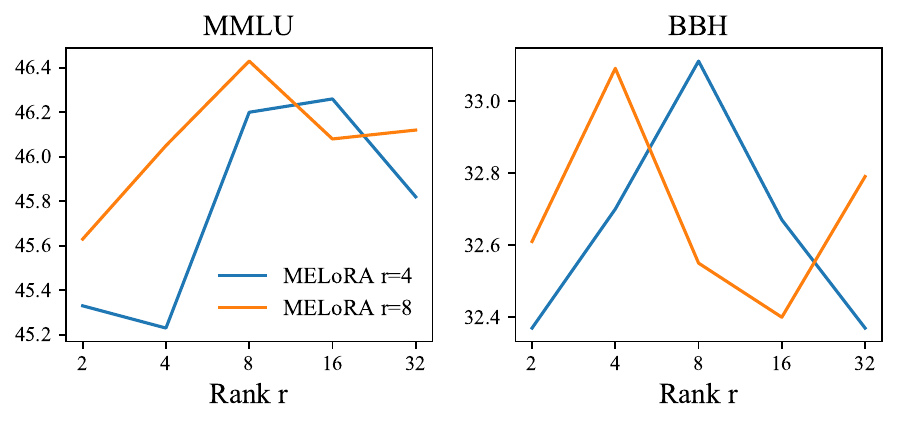}
    \caption{Performance of LoRA and \model\ with different rank $n$ and fixed $r$ on MMLU and BBH.
    }
    \label{fig:analysis_n_instruction}
\end{figure}

We report more results with different numbers of mini \acp{LoRA} in Figure~\ref{fig:analysis_n_appendix} and \ref{fig:analysis_n_instruction}.
The performance of \model\ still exhibits a pattern of initially increasing with $n$ and then decreasing, regardless of the values of $r$ on the QQP and RTE datasets.
But on the MRPC and MNLI datasets, the optimal $n$ is 1.
That is because those two datasets prefer lower ranks.
In that case, \model\ degrade to LoRA when $n=1$. 
On INSTRUCTEVAL, the performance is consistent with that in Figure~\ref{fig:analysis_n}.
The best equivalent ranks of MMLU and BBH are 64 and 32, respectively.
To achieve a specific equivalent rank, smaller the value of $r$ is, larger the value of $n$ should be.
That is consistent with Figure~\ref{fig:analysis_n} as well.

\section{Analysis of the Rank of mini \acp{LoRA}}
\label{sec:appendix_r}
\begin{figure}[ht]
    \centering
    \includegraphics[width=\linewidth]{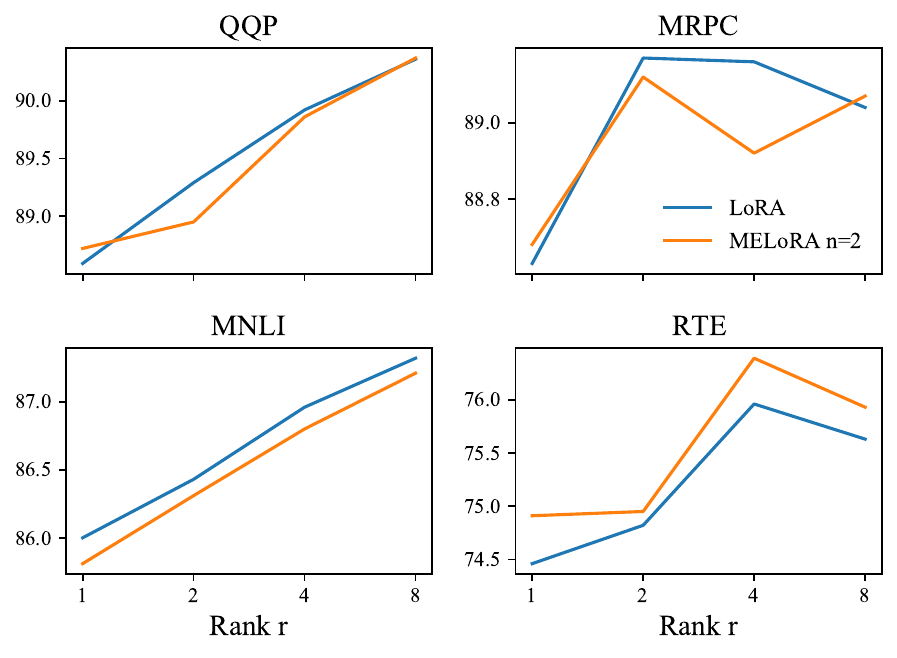}
    \caption{Performance of LoRA and \model\ with different rank $r$ and fixed $n$ on GLUE.
    }
    \label{fig:analysis_r_appendix}
\end{figure}

We report the results of different ranks of mini \acp{LoRA} on the rest datasets of GLUE in Figure~\ref{fig:analysis_r_appendix}.
\model{} outperforms LoRA, and the performance first improves and then stabilizes as $r$ increases, which is consistent with Figure~\ref{fig:analysis_r}.
Because QQP and MNLI have more training samples, the optical rank is higher.
That also proves that \model\ is more powerful with the same number of trainable parameters.

\end{document}